\pdfoutput=1

\documentclass[11pt]{article}

\usepackage{acl}

\usepackage{times}
\usepackage{latexsym}

\usepackage[T1]{fontenc}

\usepackage[utf8]{inputenc}

\usepackage{microtype}

\usepackage{inconsolata}

\usepackage{graphicx}
\usepackage[most]{tcolorbox}
\usepackage{booktabs}
\usepackage{colortbl}
\usepackage{authblk}

\title{\texttt{MAG-V}: A \underline{M}ulti-\underline{A}gent Framework for Synthetic Data \underline{G}eneration and \underline{V}erification}

\author{
  \textbf{Saptarshi Sengupta \textsuperscript{1}},
  \textbf{Harsh Vashistha \textsuperscript{1}},
  \textbf{Kristal Curtis \textsuperscript{1}},
  \textbf{Akshay Mallipeddi \textsuperscript{1}}, \\
  \textbf{Abhinav Mathur \textsuperscript{1}}, 
  \textbf{Joseph Ross\textsuperscript{1}},
  \textbf{Liang Gou\textsuperscript{1}}
\\
  \textsuperscript{1}Splunk Inc\\
  \{
  \texttt{ssengupta, hvashistha, kcurtis, amallipeddi, abhinavmathur, josephr, lgou}
  \}\texttt{@splunk.com} \\

}

\begin{document}
\maketitle
\begin{abstract}
Extending the capabilities of Large Language Models (LLMs) with functions or \textit{tools} for environment interaction has led to the emergence of the \textit{agent} paradigm. In industry, training an LLM is not always feasible because of the scarcity of domain data, legal holds on proprietary customer data, rapidly changing business requirements, and the need to prototype new assistants. Agents provide an elegant solution to the above by relying on the zero-shot reasoning abilities of the underlying LLM and utilizing tools to explore and reason over customer data and respond to user requests. However, there are two concerns here: (I) acquiring large-scale customer queries for agent testing is time-consuming, and (II) high reliance on the tool call sequence (or \textit{trajectory}) followed by the agent to respond to user queries may lead to unexpected or incorrect behavior. To address this, we propose \texttt{MAG-V}, a multi-agent framework to first generate a dataset of questions that mimic customer queries; and second, reverse-engineer alternate questions from the responses for \textit{trajectory verification}. Initial results indicate that our synthetic data can improve agent performance on actual customer queries. Furthermore, our trajectory verification methodology, inspired by \textit{distant supervision} and using traditional machine learning (ML) models, outperforms a GPT-4o judge baseline by 11\% accuracy and matches the performance of a GPT-4 judge on our constructed dataset. Overall, our approach is a step towards unifying diverse task agents into a cohesive framework for achieving an aligned objective.
\end{abstract}

\section{Introduction}

Recent advances in generative-text modeling \cite{minaee2024large, zhao2023survey} have enabled \textit{agents} \cite{xi2023rise, wang2024survey}, i.e., AI applications that use LLMs for planning and reasoning and leverage external functions (\textit{tools}) \cite{schick2024toolformer} to solve complex tasks and promote improved user interaction. Pushing the boundary even further are \textit{multi-agent} systems \cite{ijcai2024p890} that synergize the communication between multiple agents to achieve a common goal by distributing responsibilities across agents.

Building intelligent assistants \footnote{\textit{Agent} and \textit{Assistant} is used interchangeably in this paper to refer to a customer-facing LLM helper system.} to provide better customer experiences sits at the heart of every technology-driven enterprise. However, training (even fine-tuning) LLMs per customer is often infeasible due to insufficient data and privacy concerns.
This creates the perfect opportunity for using agents which can reason over private data by relying only on the zero-shot abilities of its underlying LLM.

Before deploying custom assistants, a test suite of queries representative of actual customer questions is needed. While asking customers to interact with the assistant is an option that directly provides gold data, it is time consuming and thus, does not scale well. -- \textcolor{blue}{Issue (I)}

When addressing a question, an agent decides if it can be answered using the LLM's internal knowledge or, requires information from the external environment. For the latter case, the agent invokes \textit{tools} \cite{patil2024gorilla}, i.e., functions performing dedicated tasks such as \textit{retrieving weather}, \textit{converting currency}, etc.~to gather the necessary world knowledge. The sequence of tools that an agent \textit{calls}\footnote{In agent jargon, a \textit{call} means producing a structured JSON string that can be parsed for use with the necessary function.} is referred to as its \textit{trajectory} ($T$), i.e., $T_Q = [t_1, ..., t_n]$, where $T_Q$ is the trajectory given question $Q$ and $t_i$ are individual tool calls. A tool call $t_i$ consists of two parts, viz., the tool name and the arguments. For example, in the call \texttt{get\_current\_weather(city = ``Boston'')} the tool name is \texttt{get\_current\_weather} and the argument is \texttt{city = ``Boston''}.

Determining the correctness of the agent trajectory is a non-trivial problem and forms the first part of response verification. While using strong LLMs-\textit{as-a-judge} \cite{zheng2023judging} has been proposed, the approach faces drawbacks \cite{gu2024surveyllmasajudge} such as LLM sensitivity to the input prompt \cite{anagnostidissusceptible} and inconsistent behavior of API (Application Programming Interface)-based models \cite{10.1145/3697010}; while the latter is 
mitigated by altering generation temperature, it is not completely eliminated. -- \textcolor{blue}{Issue (II)}

To address both issues, we propose \texttt{MAG-V}, a multi-agent framework for generating questions mimicking customer queries and verifying trajectories deterministically, i.e., without using LLMs to provide feedback. For verification, we take inspiration from classical ML approaches such as \textit{distant supervision} \cite{qin-etal-2018-robust} and discriminative models such as Support Vector Machines (SVM), etc., and combine them with recent advances in LLM response verification such as Self-Verification Prompting \cite{weng-etal-2023-large}. 

Overall, our contribution is three-fold:

\begin{enumerate}
    \vspace{-0.2cm}\item We propose using agents for creating synthetic data aligned with specific requirements.
    \vspace{-0.2cm}\item We introduce a deterministic method for verifying agent trajectories that does not rely on using LLMs for feedback.
    \vspace{-0.2cm}\item We show that simple ML baselines with feature engineering can match the performance of more expensive and capable models.
\end{enumerate}

\section{Related Work}

We discuss two categories of recent papers that are comparable to our work.

\noindent \textbf{Data Generation with Multi-Agent Systems} Synthetic data generation \cite{long-etal-2024-llms, chen-etal-2024-spiral} using LLMs has become a standard practice to address data scarcity and has been utilized for various tasks as inductive reasoning \cite{shao2024case2code} and question answering \cite{namboori2023gemquad}, the latter also utilizing ICL. However, using LLM agents for more complex data generation is a nascent and active area of study. \citet{DBLP:journals/corr/abs-2408-08688} utilize multiple agents (2 for generation and at most 3 for evaluation) to create a preference alignment \cite{wang2023aligninglargelanguagemodels, shen2023large} dataset. The issue with their strategy is the high reliance on LLM-as-a-judge, which we want to avoid, and no human-in-the-loop to gauge the quality of generations. The study by \citet{abdullin-etal-2023-synthetic} is perhaps the most related to ours. They utilize a two-agent system to produce a conversation dataset for solving linear programming word problems and also rely on human evaluators for feedback. Our approach differs from theirs as we require ``questions'' and not dialog data, and we prioritize tool-calling to that end. Finally, \texttt{AgentInstruct} \cite{mitra2024agentinstruct} and \citet{ge2024scaling} test the limits of multi-agent data generation. The former uses a minimum of 29 agents to transform the seed to synthetic data while the latter creates ``personas'' to apply specific transformations to the seed data, creating 25M and 1B samples respectively. Although impressive, these methods are still experimental, incur high costs, potentially introduce noise at such scales, and are unsuitable for domains such as ours due to large seed data requirements. 

\noindent \textbf{Trajectory Verification} Evaluating LLM response is an active area of study \cite{10.1145/3641289, chiang-lee-2023-large} as there is no standardized method yet for gauging their correctness. Early attempts include LLM-as-a-judge like frameworks such as \texttt{AlpacaEval} \cite{alpaca_eval, dubois2024length}, which measures a model's \textit{win rate} (WR). WR is the number of times a judge model prefers its response over a reference by considering factors such as consistency, relevancy, etc. Extending this idea for trajectory verification is \citet{qin2024toolllm} who propose using a judge model (GPT-4) to determine WR and \textit{pass rate}, i.e., whether a model is able to reach the goal in a certain number of steps. The drawback with \citet{qin2024toolllm} is their over reliance on LLM-as-a-judge for trajectory evaluation. The closest study to ours for trajectory verification is \citet{chen-etal-2024-eval} which builds on \citet{qin2024toolllm} by performing step-by-step tool call evaluation. Although they compute similar metrics as us, they ultimately incorporate LLM-based evaluation which is orthogonal to our objective.

\vspace{-5pt}
\section{Methodology}
\vspace{-5pt}

\begin{figure*}
    \centering
    \includegraphics[scale=0.85]{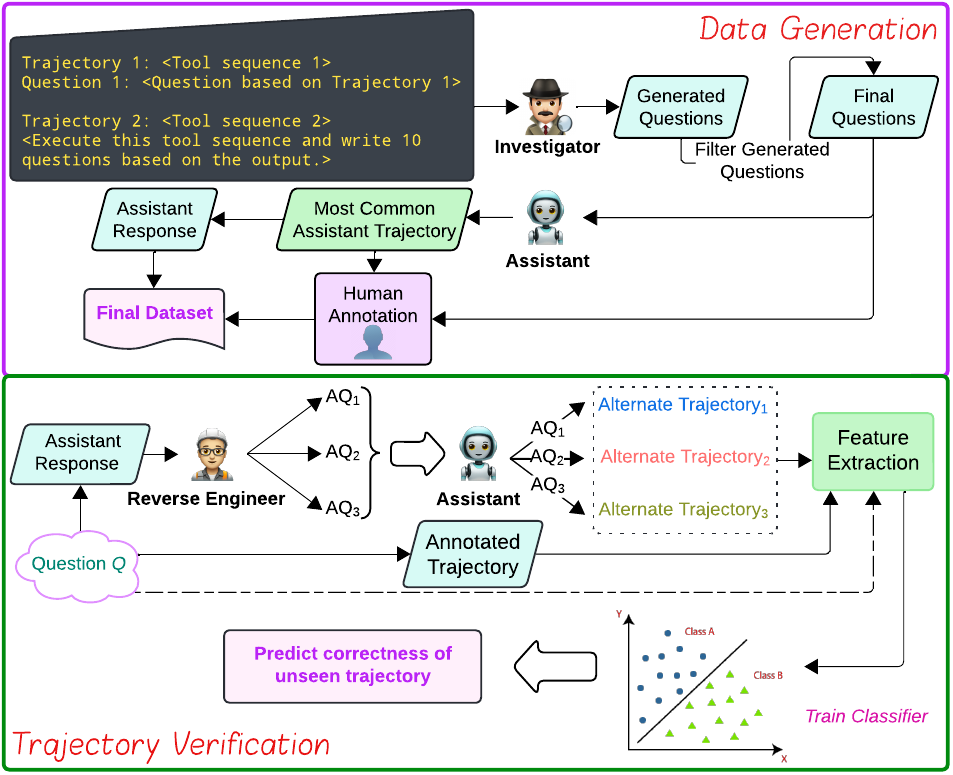}
    \caption{Overview of ~\texttt{MAG-V} [AQ = Alternate Question.]}
    \label{fig:mag-v}
    \vspace{-15pt}
\end{figure*}

Here we deconstruct the inner workings of \texttt{MAG-V} (Fig.~\ref{fig:mag-v}) starting with the dataset construction (\S \ref{sec:DG}), followed by trajectory verification (\S \ref{sec:TV}). In our experimental setup, we have three agents: \textit{investigator} (responsible for query generation), \textit{assistant} (responsible for answering queries) and \textit{reverse engineer} (responsible for creating questions based on a given response). The investigator and reverse-engineer agents use GPT-4o-mini \cite{openai2024gpt4}. Assistant uses GPT-4o-mini during data-generation and GPT-4o \cite{openai2024hello} during verification as we found 4o-mini to time-out for many reverse queries (cf.~Sec.~\ref{sec:TV}). The choice of model for each agent in our experimental setup is independent of the actual models used in our production system. Each was selected to facilitate our experimentation.

\subsection{Data Generation}
\label{sec:DG}

The requirement of the synthetic generation phase is to create a dataset of queries indicative of what a customer might ask our assistant. We begin with a small set of 19 questions written by one of our engineers to meet certain product requirements. These questions, along with their verified trajectories, form our seed dataset from which we generate synthetic samples.

Ideally, we aim to generate questions that stress test our assistant, i.e., invoke a variety of tools to handle complex user queries. To that end, we perform \textit{In-Context Learning} (ICL) \cite{dong-etal-2024-survey} (predictions based on the given examples without parameter update) to illustrate to the agent the types of trajectories and questions we desire. Each ICL sample has a trajectory $T_1$, a question that was answered using that trajectory $Q_1$, and another trajectory $T_2$ randomly sampled from the seed dataset. The \textit{investigator} executes $T_2$ and is tasked with writing 10 questions similar in style to $Q_1$ by studying the tool responses. This gives us a total of 190 synthetic questions, which we filter (cf.~Appendix \ref{sec:app_filter}) down to 45 questions.

To account for fluctuations in API calls \cite{10.1145/3697010}, we run each of the 45 questions 5 times via the assistant and consider the most common trajectory (MCT) across all trials for annotation. A response using the MCT was randomly sampled and added to the dataset.

Finally, two independent domain experts were asked to annotate the dataset. Given a question and its corresponding trajectory, the annotators needed to mark it as $0/1$, i.e., 0: Incorrect - the trajectory makes incorrect calls/makes assumptions beyond the question requirements, etc.; and 1: Correct - All steps make sense and the overall trajectory takes into account all of the user requirements. This essentially casts the problem as a binary classification. The annotators had 0.42 Cohen's Kappa indicating moderate agreement which shows the difficulty of such esoteric domains as ours\footnote{We will release details of our platform upon publication.}. The annotators agreed on 34/45 cases and disagreements were broken with the help of another expert which ultimately yields a dataset of 30 incorrect (0) and 15 correct (1) trajectories.

\subsection{Trajectory Verification}
\label{sec:TV}

The second phase of \texttt{MAG-V} is trajectory verification, i.e., ascertaining whether the assistant calls the correct sequence of tools to reach the response. This forms the first part of response verification\footnote{Verifying the assistant's answer is beyond the scope of this study.}. Our methodology is based on the hypothesis that, 

\begin{quote}
    \textit{If questions similar to a base question follow the same trajectory, we can establish a degree of confidence in the assistant's reasoning pathway. Else, high trajectory variance equates to lower confidence.}
\end{quote}

This hypothesis is inspired by two ideas, viz.,

\begin{enumerate}

    \item \textbf{Distant Supervision}: These include methods to make predictions about unlabeled data using auxiliary knowledge sources. For example, when building a relation extraction model \cite{10.1145/3241741}, we might use \textit{knowledge graphs} to make relationship inferences on unstructured text. In a related fashion, we use the trajectories of the similar questions to make predictions about our annotated trajectories.
    
    \item \textbf{Backward Reasoning}: Two recent studies, \textit{reverse question answering} (RQA) \cite{balepur2024reverse} and self-verification prompting \cite{weng-etal-2023-large} explore the idea of verifying LLM-responses by generating questions from the final response and asking the LLM to answer them. RQA is a \textit{jeopardy}-like \cite{ferrucci2013watson} technique where the generated question's answer must be the response, while the latter generates questions whose 
    answers contain all facts entailed in the original question. Inspired by these methods, we adopt a similar backward question generation strategy, but we study trajectory verification instead of response.
\end{enumerate}

To verify the MCT of a question $Q$, we first ask the reverse engineer agent to create 3 \textit{alternate} questions (AQ) based on the assistant's response using the MCT. The agent was instructed to write questions that captured all of the main points of the response. The assistant then answers the AQ's which leads to 3 alternate trajectories (AT).

Using the annotated (or base) trajectory (BT) and the AT's, we extract \textit{features}\footnote{Ideally, we should refer to these as \textit{metrics}. However, we use \textit{features} to distinguish between evaluation metrics and values used to train our models.} to determine the similarity between them. These features are a combination of statistical and embedding-based measures. In total, we compute six features between the BT and AT's (cf.~Appendix \ref{sec:Traj_Features}), viz., \textbf{EM} (Exact Match - 0/1 measure of equality); \textbf{EDIT} (minimum number of string edits needed to convert AT to BT); \textbf{GEDIT} (Graph Edit Distance between the trajectories formulated as graphs); \textbf{SS} (Semantic (Cosine) Similarity between BT and AT); \textbf{AO} (Argument Overlap - count of common arguments between BT and AT) and \textbf{LCSS} (Longest Common Sequence from Starting Call to measure the extent of commonality between BT and AT from a common point). Additionally, we also extract \textbf{TF-IDF} (Term Frequency-Inverse Document Frequency) features from the base questions to provide grounding context to the models as a trajectory can only be analyzed in relation to its question.

Finally, using these features we train discriminative ML models using stratified ten-fold cross-validation to account for label imbalance. Overall, we test 7 ML models (Random Forest, Logistic Regression, Naive Bayes, $k$-Nearest Neighbours ($k$-NN), Support Vector Machines (SVM), Decision Tree, XGBoost) and report the mean accuracy and F1 across all folds and 3 random seeds. Apart from AO, we test the influence of including and excluding the tool arguments for each measure for trajectory verification. All models used default settings except for $k$-NN which we found to have the best performance at $k=5$ (without arguments verification) and $k=4$ (with arguments verification).

\subsubsection{GPT-baseline}
To compare against a strong LLM-as-a-judge, we selected GPT-4 \cite{openai2024gpt4technicalreport} and GPT-4o. Each model was tested on the same test splits and random seeds from the cross-validation to provide consistent evaluation. We provided the same annotation rubric to the models, but switched the position of the labels to account for \textit{recency bias} \cite{pmlr-v139-zhao21c}, i.e., the tendency of LLMs to repeat the last seen values in the prompt. Before generating its 
score, the LLM was asked to provide a rationale as it has been shown \cite{wang-etal-2024-large-language-models-fair} to lead to better evaluation. Additionally, we provided one example each of a correct and incorrect $Q + T$ pair from the seed dataset to further help the models. Finally, we also tested
performance with and without using the default system prompt (\textit{You are a helpful AI Assistant.}) in \texttt{Autogen} \cite{wu2024autogen}, the framework used to create our agents. All other model parameters were set to their default.

\section{Experiments}

In this section, we first explain the benefit of generating synthetic questions and then detail the results from our trajectory verification trials.

\subsection{Utilizing Synthetic Data}

The goal of creating synthetic questions was to simulate user queries to test the capabilities of our assistant. As a by-product of this step, we also find that the generated data can be used to aid our assistant in fixing its mistakes, and answering questions better through ICL. The following is an example of a \textbf{real customer query} (ignore the grammar issue), modified to work in our test environment.

\begin{tcolorbox}[colback=red!5,colframe=red!50!black]
I see that the paymentservice on production had an increased error rate around 7:59pm EST 10/10/2024, can you dive into that service during that time and a half hour and each side and find out what the most common errors associated were?
\end{tcolorbox}

The assistant usually runs GPT-4o as its backbone.~However, to test for cost-efficiency, we also considered 
4o-mini.~To answer this question, the assistant needs to use the correct time range, which is 7:29 PM to 8:29 PM on October 10.~We run the above query with both 4o and 4o-mini, and also add all 45 synthetic questions and their MCT as ICL samples.~The 4o-backed assistant with and without the samples fails to use the correct time range: 
it uses \texttt{\{``time\_range'':\{``start'':``2024-10-10T23:29\\:00Z'', ``stop'':``2024-10-11T00:29:00Z''\}\}}. While the start date is correct, the start/end times and end date are wrong. With 4o-mini, the assistant spirals into a chain of retries, unable to fix itself even after 10 attempts. However, by adding our generated questions and their MCT as ICL samples, the assistant was able to 
to figure out the correct time range in the first try, \texttt{\{``time\_range'': \{``start'': ``2024-10-10T19:29:00Z'', ``stop'': ``2024-10-10T20:29:00Z''\}\}\}}.~This signals potential cost-benefits, i.e., by coupling a cheaper model with ICL samples, we can guide it to perform better or on-par with more expensive LLMs.

\begin{figure*}
    \centering
    \includegraphics[width=\textwidth]{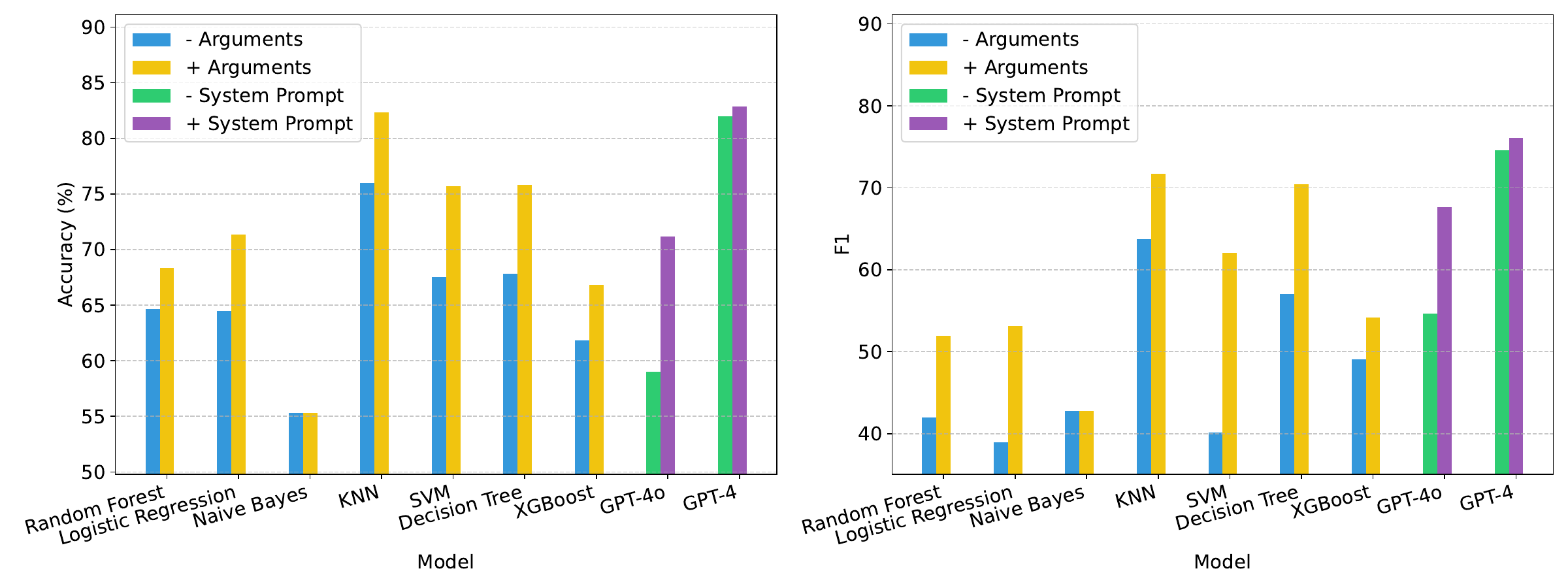}
    \caption{Accuracy (left) and F1 (right) scores from all ML models using all features v/s GPT-baselines.}
    \label{fig:TV_Full}
    \vspace{-12pt}
\end{figure*}

\subsection{Trajectory Verification}

The results from our verification trials (accuracy and F1) are shown in Figure \ref{fig:TV_Full}. As we see, each ML model benefits from features that consider tool arguments. The reason for experimenting with removing the arguments is to see if the \textit{overall tool calls} were consistent across the base and alternate trajectories. However, it makes sense that by considering the arguments, the models perform better as they have more fine-grained features to learn from. Furthermore, by excluding the arguments, the boundary separating the correct and incorrect trajectories becomes less prominent since many trajectories with the same calls get placed in both classes. For example, if $T_1 = T_2 = [t_1, t_2, t_3]$, where $T_1$ is incorrect and $T_2$ is correct for their respective questions, the model learns the same features for each class which in turn distorts its performance. By adding the arguments, the model is able to learn a better decision boundary. Overall, we see that our $k$-NN model shows the best accuracy and F1 scores across all ML models.

Scores from the GPT-models are a bit surprising. Firstly, GPT-4o, which is claimed to be a much stronger model than GPT-4 \cite{openai2024hello}, is outperformed by the latter by quite a margin. Second, by simply adding the system prompt (\textit{You are a helpful AI Assistant.}), the accuracy of GPT-4o increases by about 12\%, which hearkens to our point on prompt sensitivity, i.e., making such small modifications to a prompt can have tremendous impact on the output. Finally, our \textbf{$k$-NN model shows 11\% accuracy and 4\%  F1 improvement over GPT-4o} which demonstrates the effectiveness of such a simple model over a more complex LLM. While our $k$-NN model (82.33\%) matches GPT-4 in accuracy (82.83\%), it lags behind in F1 (71.73 vs.~76.06), indicating a reduced alignment of predictions.~We believe that this can be addressed if we have a larger dataset for the model to learn from.

We also consider feature ablations for our models shown in Table \ref{tab:features_ablation}. These trials only considered the six trajectory features (with arguments) to help us understand the extent of performance we can get by only conditioning our models on the trajectories. As the number of features here is much lower than the standard setup, we run five-fold cross-validation (to avoid overfitting) but use the same random seeds as before. We test each feature combination (by considering 1, $\ldots$, 6 features at a time) and report scores for those models which had the best score using the same feature subset across each seed. The GPT-baselines shown in Table \ref{tab:features_ablation} are their best scores, i.e., using the system prompt. In this setup, we see that our best models come close to GPT-4 and outperform GPT-4o in accuracy. Additionally, the distribution of predictions (F1) is much more aligned with that of GPT-4. Interestingly the models utilize only a single feature, i.e., EDIT distance, to make their best predictions. We reason that this is because EDIT is a more lenient metric than the others, as it determines the amount of work needed to align two trajectories, rather than penalizing mismatches which in turn provides more signal to the models.

Finally, we discuss the trade-offs with our approach versus using LLM-as-a-judge. At the outset, we see that performance for both are comparable using either question and trajectory features, or just the latter in isolation. However, as our method requires three alternate questions and consequent trajectories to be generated, it incurs more API calls. At scale, this might accrue cost if using an expensive LLM such as GPT-4 or GPT-4o. However, switching to a cheaper model such as GPT-4o-mini or open-source models such as Llama 3 \cite{dubey2024llama} can alleviate this to a great extent. Where our approach wins over the GPT benchmark is in \textit{determinism}, i.e., for the exact same conditions, it will always give the same result, which cannot be said for API-based LLMs. However, it might be argued that changes in the alternate trajectories can impact the features being learned. To this, we explain that our approach follows a \textit{best-of-N-predictions} strategy, i.e., we base our decisions on what the majority of the trajectories predict as opposed to a single trajectory. This leads to a model that is more robust and confident in its predictions. 

\begin{table}[ht]
\centering
\resizebox{\columnwidth}{!}{%
\begin{tabular}{@{}cccc@{}}
\toprule
Model          & Feature(s)                   & Accuracy (\%)  & F1             \\ \midrule
GPT-4o         & -                            & 71.11          & 69.85          \\
\textbf{GPT-4} & -                            & \textbf{82.22} & \textbf{77.94} \\ \midrule
\textbf{Random Forest} &
  \cellcolor[HTML]{FFFFFF}\textbf{EDIT} &
  \cellcolor[HTML]{FFFFFF}\textbf{80} &
  \cellcolor[HTML]{FFFFFF}{\color[HTML]{1F1F1F} \textbf{75.3}} \\
Decision Tree  & \cellcolor[HTML]{FFFFFF}EDIT & 77.04          & 70.87          \\
\textbf{XGBoost} &
  \cellcolor[HTML]{FFFFFF}\textbf{EDIT} &
  \cellcolor[HTML]{FFFFFF}{\color[HTML]{1F1F1F} \textbf{80}} &
  \cellcolor[HTML]{FFFFFF}\textbf{75.3} \\ \bottomrule
\end{tabular}%
}
\caption{ML models trained using the best trajectory feature subset vs.~GPT-baselines. Using EDIT distance, each model displays the best accuracy and F1 across all random seeds. Each score is the average from all trials.}
\label{tab:features_ablation}
\vspace{-12pt}
\end{table}

\section{Conclusion and Future Work}

We introduce \texttt{MAG-V}, a framework for generating synthetic questions using LLM agents and verifying the trajectory taken by an agent to answer questions. The novelty of our verification system lies in its deterministic aspect, i.e., reducing the reliance on API-based LLMs to judge the veracity of results (trajectories). That said, we believe there is more work to be done. First, we need to test how our method scales with the 
number of samples. Second, we use TF-IDF features to represent questions. We will look at ways to better ground trajectories to their associated questions to improve verification performance. Finally, during annotation, we observed that some ``incorrect'' trajectories \textit{can} be marked as correct, considering the complexity of the questions. As such, we aim to smooth our labels to have 3-classes (correct, \textit{partially correct} and incorrect). This will be a bigger challenge for the GPT models as they have been shown \cite{10479409} to struggle with multiclass classification. We aim to see how our models fare 
under such a setup.

\bibliography{acl_latex}

\appendix

\section{Filtering Generated Questions}
\label{sec:app_filter}

To avoid having questions of the same type for our test set, we filter the generated questions. We start by embedding the questions using  BAAI General Embeddings (\texttt{BGE}) \cite{10.1145/3626772.3657878} as they achieve state-of-the-art performance on MTEB (Massive Text Embedding Benchmark) \cite{muennighoff-etal-2023-mteb}. This yields $R^{1024}$ question embeddings. Using UMAP (Uniform Manifold Approximation and Projection) \cite{mcinnes2018umap-software} for dimension-reduction to $R^2$, we cluster the resulting embeddings with HDBSCAN \cite{10.1007/978-3-642-37456-2_14}. This results in 11 clusters which we manually verified as being sufficiently diverse in their themes such as requesting information on a given service in our environments, issues related to a hardware component, etc. Considering the mean of all embeddings in a cluster as the centroid, we take the top-5 questions with the least distance to the centroid, which gives a total of 55 questions. Finally, we remove 10 questions that are either generic and do not require tool calls, or pertain to an aspect of our platform that is under active development. This yields a dataset of 45 questions.

\section{Trajectory Features}
\label{sec:Traj_Features}

We extract six features across the base and alternate trajectories, described as follows:

\begin{itemize}
    \item \textbf{EM} (Exact Match): A binary measure of equality, i.e., $EM = 1$, if $BT = AT$ verbatim, else 0.

    \item \textbf{EDIT}: Levenshtein Edit Distance \cite{10.5555/1822502} between BT and AT, i.e., the minimum number of \textit{edits} (add, delete, substitute) required to modify the AT to become the BT.

    \item \textbf{GEDIT} (Graph Edit Distance): As a trajectory is a sequence of operations, it can also be viewed as a \textit{directed graph}. As such, we decide to measure the edit distance required to make the BT and AT graphs isomorphic, i.e., similar in structure. GEDIT is similar to EDIT, but operates on nodes and edges. 

    \item \textbf{SS} (Semantic Similarity): This measures cosine similarity between the BT and AT embedded using \texttt{BGE-Large} embeddings.

    \item  \textbf{AO} (Argument Overlap): Count of arguments common to BT and AT for each tool call pair. AO is an F1 score ($2PR/(P+R)$), where:

    \begin{itemize}
        \item \textit{overlap} ($O$) = number of common arguments across all calls for BT and AT.
        \item \textit{precision} ($P$) = $O$ / total number of arguments in AT.
        \item \textit{recall} ($R$) = $O$ / total number of arguments in BT.
    \end{itemize}

    \item \textbf{LCSS} (Longest Common Sequence from Starting Call): The longest tool call sequence common to BT and AT starting from the first call. Similar to AO, LCSS is an F1 score with: 

    \begin{itemize}
        \item $L$ = Length of longest sequence across BT and AT from the starting call.
        \item $P$ = $L$ / total number of calls in AT.
        \item $R$ = $L$ / total number of calls in BT.
    \end{itemize}

\end{itemize}

\end{document}